%% file: main.tex
\documentclass[10pt,twocolumn,letterpaper]{article}

\usepackage[pagenumbers]{cvpr}
\input{preamble}
\usepackage{amsmath, amssymb, bm, booktabs,tabularx,array}
\definecolor{cvprblue}{rgb}{0.21,0.49,0.74}
\usepackage[breaklinks,colorlinks,allcolors=cvprblue]{hyperref}

\title{DuCo: Exploiting Dual Cross-Layer Consistency for Sparse Geometric Attention}

\author{Liangyu Xiang\textsuperscript{1,2} \qquad Junyu Gao\textsuperscript{1,2\textdagger} \qquad Xiaofeng Wang\textsuperscript{4} \qquad Changsheng Xu\textsuperscript{1,2,3}\\
\textsuperscript{1}State Key Laboratory of Multimodal Artificial Intelligence Systems (MAIS),\\
Institute of Automation, Chinese Academy of Sciences (CASIA)\\
\textsuperscript{2}School of Artificial Intelligence, University of Chinese Academy of Sciences (UCAS)\\
\textsuperscript{3}Peng Cheng Laboratory, Shenzhen, China \qquad \textsuperscript{4}GigaAI\\
{\tt\small xiangliangyu2023@ia.ac.cn; \{junyu.gao, csxu\}@nlpr.ia.ac.cn; wangxiaofeng2020@ia.ac.cn}
}

\begin{document}
\maketitle
\renewcommand{\thefootnote}{\fnsymbol{footnote}}
\footnotetext[2]{Corresponding author.}
\renewcommand{\thefootnote}{\arabic{footnote}}
\input{sec/0_abstract}

\input{sec/1_intro.tex}

\input{sec/2_related_work.tex}

\input{sec/3_method.tex}

\input{sec/4_experiements.tex}

% Flush experimental floats before the conclusion.
\clearpage
\input{sec/conclusion.tex}
{
    \small
    \bibliographystyle{ieeenat_fullname}
    \bibliography{main}
}

\end{document}

%% file: preamble.tex
\usepackage{multirow}
\usepackage{makecell}
\usepackage{colortbl}

%% file: sec/0_abstract.tex
\begin{abstract}
Dense global attention is crucial for capturing cross-view geometry in visual geometry foundation models, yet its quadratic complexity limits efficient inference on long image sequences. Existing sparse attention methods select entries of the global attention matrix guided by attention probabilities and construct masks independently across layers, overlooking value redundancy and cross-layer information reuse. By analyzing attention distributions and clustering value representations across layers, we reveal \emph{dual cross-layer consistency}, characterized by highly similar attention probability distributions and localized changes in value redundancy structures. Accordingly, we propose \textbf{DuCo}, a training-free sparse attention framework that exploits historical attention information and value redundancy for efficient geometric inference. Specifically, DuCo leverages historical attention information to estimate current-layer dense attention probabilities and introduces value clustering to exploit redundancy, with clusters refined through local updates across layers. Both factors are integrated through probability-weighted value distortion to jointly optimize four cluster representation states under an attention budget for sparse mask construction. Experiments across three visual geometry foundation models and five benchmarks demonstrate favorable accuracy--efficiency trade-offs.
\end{abstract}

%% file: sec/1_intro.tex
\section{Introduction}
\label{sec:intro}

\begin{figure}[t]
    \centering
    \includegraphics[width=\columnwidth]{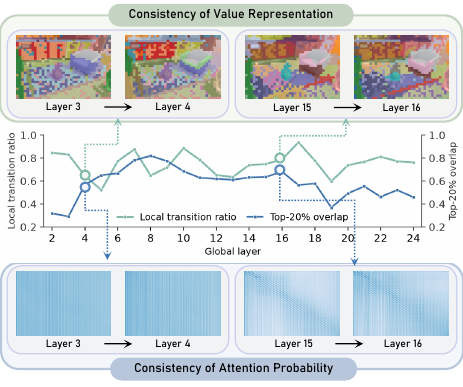}
    \caption{Cross-layer consistency of global attention layers in VGGT~\cite{wang2025vggt}. Top-20\% overlap measures the proportion of high-probability attention entries shared between consecutive layers, while local transition ratio measures the proportion of changed value cluster assignments that transition to neighboring clusters. Visualizations at Layers 3--4 and 15--16 provide representative cross-layer comparisons of attention and value clustering.}
\label{fig:cross_layer_persistence}
    \vspace{-5mm}
\end{figure}

Recovering 3D scene geometry from images is a fundamental problem in computer vision, providing the spatial understanding required for autonomous navigation, robotic perception, and augmented reality~\cite{schonberger2016structure,mur2017orb,klein2007parallel}. Recent advances in feed-forward visual geometry foundation models, such as VGGT~\cite{wang2025vggt} and $\pi^3$~\cite{wang2026pi}, have enabled joint estimation of camera poses and dense scene geometry from uncalibrated multi-view images in a single forward pass. Central to these models is global self-attention, which integrates information across input views to capture geometric relationships. However, this mechanism densely evaluates all entries of the global attention matrix, incurring a computational cost that grows quadratically with the total token count. As the number of input views increases, the computational cost of dense attention becomes a major bottleneck to efficient geometric inference.

Existing sparse attention methods for visual geometry foundation models reduce computational costs by selecting important entries from the global attention matrix or adapting sparsity across layers and heads~\cite{wang2026block,deng2026saf3r,kim2026hess}. However, these methods primarily emphasize attention importance, without fully accounting for redundancy among the associated value representations. Since dense attention computes a probability-weighted sum of values, even high-probability entries may convey redundant information when their corresponding values are similar. Moreover, existing methods largely rely on layer-specific information for sparsification, leaving the cross-layer relationships of attention probabilities and value redundancy insufficiently exploited. These limitations motivate a joint investigation of attention probabilities and value redundancy, particularly their relationships across successive global attention layers.

To investigate these relationships, Fig.~\ref{fig:cross_layer_persistence} quantifies the evolution of attention probabilities and value redundancy across VGGT layers, using the top-20\% entry overlap and local transition ratio of cluster assignments. The results reveal dual aspects of cross-layer consistency, where attention probability distributions remain highly similar across neighboring layers, while changes in value redundancy structures are largely localized. Visualizations of attention matrices and value clustering at Layers 3--4 and 15--16 further illustrate these observations. This consistency motivates reusing historical attention information and value clustering results as priors for subsequent layers. Building on this insight, we propose a training-free sparse attention framework that propagates and refines historical priors across global attention layers. Specifically, previously computed attention logits and current-layer regional estimates are combined to approximate dense attention probabilities by accounting for unobserved entries in softmax normalization. Cross-layer regional consistency further calibrates historical probabilities, while current regional estimates provide probabilities for previously unobserved entries. Meanwhile, historical value clusters are locally refined using current-layer value representations, avoiding repeated global clustering. Based on the estimated probabilities and refined clusters, four representation states, Drop, Rep, History, and Exact, are introduced to capture varying degrees of attention importance and value redundancy. These states are jointly selected by minimizing probability-weighted value distortion under an attention budget to construct sparse attention masks. The resulting attention information and updated value clusters are propagated to subsequent layers for efficient reuse.

Our contributions are summarized as follows:
\begin{itemize}
    \item We identify two forms of cross-layer consistency in global attention, characterized by similar attention probability distributions across neighboring layers and localized changes in value redundancy structures.

    \item We develop a training-free sparse attention framework that leverages cross-layer consistency to propagate historical priors. Historical attention information guides dense attention probability estimation, while inherited value clusters are refined through local updates. Both components are further integrated into a probability-weighted value distortion objective to jointly optimize cluster representation states for sparse mask construction.
    
    \item Extensive experiments across diverse visual geometry foundation models and benchmarks demonstrate that DuCo substantially improves inference efficiency while preserving competitive geometric reconstruction accuracy, achieving a favorable accuracy--efficiency trade-off.
\end{itemize}

%% file: sec/2_related_work.tex
\section{Related Work}

% \noindent\textbf{Feed-forward Visual Geometry Transformers.} Traditional multi-view geometry pipelines rely on explicit feature matching and iterative optimization~\cite{triggs1999bundle,schonberger2016structure,mur2015orb}. Feed-forward visual geometry Transformers instead infer cross-view structure directly from image tokens, spanning pairwise point-map regression~\cite{wang2024dust3r}, joint multi-view reconstruction~\cite{yang2025fast3r,cabon2025must3r}, and unified prediction of camera poses, depth, and point maps~\cite{wang2025vggt}. These models differ in how geometric information is represented and exchanged. VGGT~\cite{wang2025vggt} combines frame-wise and global attention, while $\pi^3$~\cite{wang2026pi} employs a permutation-equivariant formulation and VGGT-$\Omega$~\cite{wang2026vggt} adopts register-based scene aggregation. Other advances explore depth-ray representations~\cite{lin2025depth} and extend geometric inference to heterogeneous inputs~\cite{keetha2026mapanything,karhade2026any4d}, streaming observations~\cite{yuan2026infinitevggt,zhuo2026streaming}, and dynamic scenes~\cite{zhou2026page,fang2026more, zhang2026efficiently}. Despite these advances, global attention remains essential for cross-view information exchange in many architectures, while its quadratic complexity increasingly limits long-sequence inference.

\noindent\textbf{Feed-forward Visual Geometry Transformers.} Traditional multi-view geometry pipelines rely on explicit feature matching and iterative optimization~\cite{triggs1999bundle,schonberger2016structure,mur2015orb}. Feed-forward visual geometry Transformers instead infer cross-view structure directly from image tokens, spanning pairwise point-map regression~\cite{wang2024dust3r}, joint multi-view reconstruction~\cite{yang2025fast3r,cabon2025must3r}, and unified prediction of camera poses, depth, and point maps~\cite{wang2025vggt}. Compared with pairwise reconstruction, joint multi-view formulations aggregate geometric information from multiple observations within a single forward pass, reducing the need for separate cross-view alignment. This places greater emphasis on effective feature exchange across views, particularly as the number of input images increases. These models differ in how geometric information is represented and exchanged. VGGT~\cite{wang2025vggt} combines frame-wise and global attention, while $\pi^3$~\cite{wang2026pi} employs a permutation-equivariant formulation and VGGT-$\Omega$~\cite{wang2026vggt} adopts register-based scene aggregation. Other advances explore depth-ray representations~\cite{lin2025depth} and extend geometric inference to heterogeneous inputs~\cite{keetha2026mapanything,karhade2026any4d}, streaming observations~\cite{yuan2026infinitevggt,zhuo2026streaming}, and dynamic scenes~\cite{zhou2026page,fang2026more,zhang2026efficiently}. Despite these advances, global attention remains essential for cross-view information exchange in many architectures, while its quadratic complexity increasingly limits long-sequence inference.

\noindent\textbf{Efficient Inference for Visual Geometry Transformers.} Efficient inference for visual geometry Transformers has been explored through token reduction~\cite{shen2026fastvggt,chen2026co,wang2026httm,you2026regimevggt,tang2026spark3r}, compact scene representations~\cite{elflein2026vgg,jin2026zipmap,xie2026scal3r}, and sparse attention~\cite{deng2026saf3r,wang2026block}. FastVGGT~\cite{shen2026fastvggt} prunes redundant tokens, while FlashVGGT~\cite{wang2026flashvggt} compresses global context into descriptor tokens. In contrast, sparse attention reduces the number of evaluated entries in the global attention matrix while retaining the original token sequence. Block-Sparse Global Attention~\cite{wang2026block} selects salient blocks using pooled query and key features, while SAF3R~\cite{deng2026saf3r} combines head-wise profiling with online adaptation. HeSS~\cite{kim2026hess} reallocates sparsity according to head sensitivity, whereas Speed3R~\cite{ren2026speed3r} learns coarse-to-fine sparse selection. AVGGT~\cite{sun2026avggt} instead simplifies early global layers and subsamples keys and values. Despite their different strategies, these methods primarily rely on layer-specific attention importance or sparsity characteristics, often prioritizing high-probability entries to preserve dense attention outputs. 
In the broader context of model adaptation, PhaseLoRA~\cite{guo2026phaselora} explores control-regime-conditioned low-rank adaptation for continuous-action vision-language-action policies. Beyond geometric models, prior studies have theoretically examined the benefits of considering value representations in attention approximation~\cite{gupta2021value}, while others directly reuse important token selections~\cite{zarch2026delta} or attention weights~\cite{liao2024beyond} across layers. These approaches explore value-aware approximation and historical information reuse, but do not jointly exploit their cross-layer relationships for sparsification. In contrast, DuCo leverages dual cross-layer consistency in attention probabilities and value redundancy, adapting historical information and integrating both factors through a unified probability-weighted distortion objective.

\begin{figure*}[t]
    \centering
    \includegraphics[width=\textwidth]{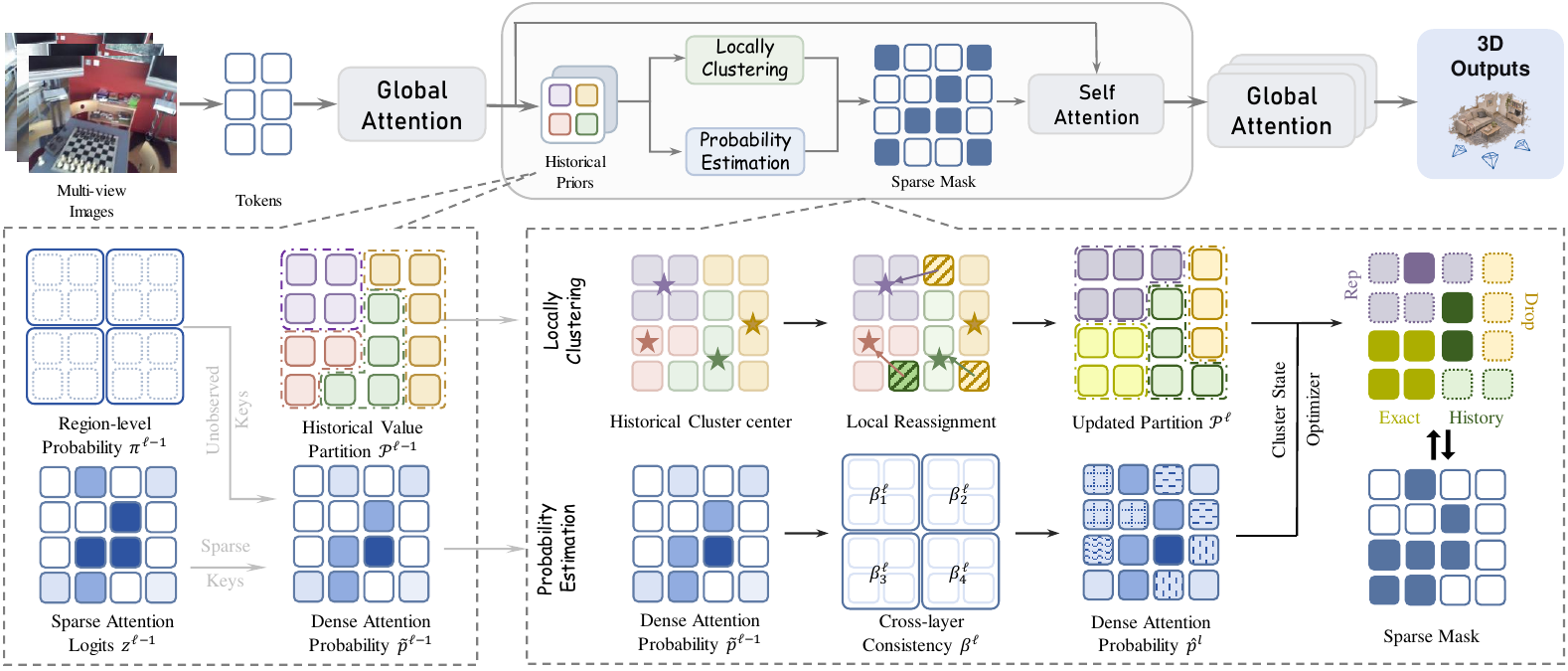}
    \caption{\textbf{Overview of DuCo.} Historical attention information and value clusters from the preceding global attention layer guide current-layer dense attention probability estimation and local cluster refinement. The resulting probabilities and clusters are jointly used to optimize cluster representation states based on probability-weighted value distortion, constructing sparse attention masks. Updated attention information and value clusters are propagated to subsequent layers as historical priors.}
    \vspace{-2mm}
    \label{fig:method_overview}
\end{figure*}

%% file: sec/3_method.tex
% Requires amsmath, amssymb, and bm (optional, if replacing \mathbf with \bm).
% Body-only section; paste into the main CVPR/ICCV manuscript.
\section{Method}
\label{sec:method}

\subsection{Cross-Layer Consistency in Global Attention}
\label{sec:cross_layer_persistence}

Existing sparse attention methods for visual geometry foundation models primarily select entries in the global attention matrix based on estimated attention probabilities independently at each layer~\cite{wang2026block,deng2026saf3r}, overlooking the influence of value representations on attention sparsification. However, dense global attention computes its output as a probability-weighted sum of value representations, $\mathbf{o}_{i}^{(\ell)} = \sum_{j=1}^{N} p_{ij}^{(\ell)} \mathbf{v}_{j}^{(\ell)}$, suggesting that value redundancy should also be considered. We therefore investigate cross-layer relationships in global attention from two perspectives: attention probability distributions and value redundancy structures, as illustrated in Fig.~\ref{fig:cross_layer_persistence}.

Fig.~\ref{fig:cross_layer_persistence} illustrates the cross-layer relationships of attention probabilities and value redundancy in VGGT. These visualizations reveal pronounced \emph{dual cross-layer consistency} in attention probability distributions and value redundancy structures. Specifically, attention distributions remain highly similar across neighboring layers, with high-probability entries for the same query largely preserved, as exemplified by Layers 15 and 16 in Fig.~\ref{fig:cross_layer_persistence}(a). This consistency may reflect the shared underlying scene geometry, which encourages attention to emphasize similar geometric correspondences despite evolving token representations. Such consistency suggests that historical attention information can guide the identification of important entries in subsequent layers, reducing redundant computations. To characterize value redundancy, we perform value clustering across layers, as shown in Fig.~\ref{fig:cross_layer_persistence}(b). Although cluster assignments evolve with network depth, their changes remain largely localized, with an average local transition ratio of 70.78\% across consecutive global attention layers of VGGT on 7Scenes. This localized variation suggests relatively stable value redundancy structures, allowing historical clustering results to be efficiently refined rather than recomputed globally at each layer. Collectively, these observations motivate propagating historical priors across layers to support attention probability estimation and value redundancy modeling.

\subsection{Propagation of Historical Priors}
\label{sec:prior_propagation}

The cross-layer consistency observed in Sec.~\ref{sec:cross_layer_persistence} suggests that previously computed attention information and value clustering results can serve as historical priors for subsequent layers. However, directly reusing these priors presents two challenges. First, sparse attention probabilities differ from their dense counterparts due to the exclusion of unobserved entries from softmax normalization, while variations in attention distributions across layers further limit the reliability of historical probabilities. Second, although value redundancy structures remain largely consistent, changes in value representations may lead to variations in cluster assignments, making direct reuse of historical clustering results inaccurate. To address these challenges, we develop two complementary approaches: \emph{history-guided attention probability estimation}, which accounts for unobserved entries and calibrates historical probabilities according to cross-layer variations, and \emph{local refinement of value clusters}, which updates inherited clustering results using current-layer value representations. Together, these approaches enable efficient propagation and adaptation of historical priors for subsequent attention sparsification.

 \paragraph{History-guided attention probability estimation.}
Although the cross-layer consistency observed in Fig.~\ref{fig:cross_layer_persistence} is based on dense attention probabilities, only sparse attention probabilities are available during sparse inference. These probabilities cannot be directly reused as historical priors, since unobserved entries are excluded from softmax normalization, introducing discrepancies from the dense attention distribution. To address this issue, we estimate the dense attention probabilities of historical entries by accounting for the contributions of unobserved keys. For query $i$ at layer $\ell$, let $\mathcal{H}_i^{(\ell-1)}$ denote the set of keys corresponding to entries evaluated in the preceding layer. We reuse their computed logits $z_{ij}^{(\ell-1)}$ and approximate the remaining logits through region-level attention. Specifically, all keys are partitioned into $R$ non-overlapping regions $\{\mathcal{R}_r\}_{r=1}^{R}$, with each region represented by its mean key feature $\bar{\mathbf{k}}_r^{(\ell-1)}$. The unobserved logits are approximated by $\bar{z}_{ir}^{(\ell-1)}=\mathbf{q}_i^{(\ell-1)\top}\bar{\mathbf{k}}_r^{(\ell-1)}/\sqrt{d}$. Accordingly, the region-level probabilities and dense attention probability estimates for historical entries are given by
\begin{equation}
\begin{gathered}
\pi_{ir}^{(\ell-1)}
=
\frac{n_r e^{\bar{z}_{ir}^{(\ell-1)}}}
{\sum_{t=1}^{R}n_t e^{\bar{z}_{it}^{(\ell-1)}}},
\\[6pt]
\tilde{p}_{ij}^{(\ell-1)}
=
\frac{e^{z_{ij}^{(\ell-1)}}}{
\displaystyle
\sum_{k\in\mathcal{H}_i^{(\ell-1)}}e^{z_{ik}^{(\ell-1)}}
+
\sum_{r=1}^{R}
(n_r-m_{ir}^{(\ell-1)})e^{\bar{z}_{ir}^{(\ell-1)}}
},
\end{gathered}
\label{eq:historical_probability}
\end{equation}
where $n_r=|\mathcal{R}_r|$ and $m_{ir}^{(\ell-1)}=|\mathcal{R}_r\cap\mathcal{H}_i^{(\ell-1)}|$. These probability estimates are then propagated to the next layer as historical attention priors, preserving regional attention information and fine-grained importance estimates for previously evaluated entries.

However, historical probability estimates may become inaccurate as attention distributions evolve across layers. To account for such variations, we retain the same region partition and compute current-layer regional probabilities $\pi_{ir}^{(\ell)}$. The consistency between consecutive layers is quantified by
\begin{equation}
\beta_{ir}^{(\ell)}
=
\left(
\frac{\min(\pi_{ir}^{(\ell)},\pi_{ir}^{(\ell-1)})}
{\max(\pi_{ir}^{(\ell)},\pi_{ir}^{(\ell-1)})}
\right)^{\tau},
\label{eq:regional_consistency}
\end{equation}
where $\tau$ controls the sensitivity to cross-layer variations. A higher $\beta_{ir}^{(\ell)}$ indicates greater regional consistency and thus stronger confidence in the historical probability estimates.

Using $\beta_{ir}^{(\ell)}$ to quantify cross-layer consistency, we estimate the current-layer dense attention probabilities of historical entries by calibrating their historical importance, while estimating those of previously unobserved entries from the current regional distribution:
\begin{equation}
\hat{p}_{ij}^{(\ell)}
=
\frac{1}{\mathcal{Z}_i^{(\ell)}}
\begin{cases}
u_{ij}^{(\ell)}
\left(
\dfrac{\tilde{p}_{ij}^{(\ell-1)}}{u_{ij}^{(\ell-1)}}
\right)^{\beta_{ir}^{(\ell)}/2},
& j\in\mathcal{H}_i^{(\ell-1)},\\[8pt]
u_{ij}^{(\ell)},
& j\notin\mathcal{H}_i^{(\ell-1)},
\end{cases}
\label{eq:probability_fusion}
\end{equation}
where $u_{ij}^{(t)}=\pi_{ir}^{(t)}/n_r$ for $j\in\mathcal{R}_r$ denotes the region-uniform probability estimate, and $\mathcal{Z}_i^{(\ell)}$ is the normalization constant over all keys.
The weights are normalized over all keys to obtain $\hat{p}_{ij}^{(\ell)}$. This formulation incorporates historical probabilities when regional attention remains consistent, while attenuating its influence under cross-layer variations. Unobserved entries are estimated from the current regional prior, enabling dense attention probability estimation without evaluating the full attention matrix.

\paragraph{Local refinement of value clusters.}
Similar to attention distributions, value representations evolve across successive layers, despite the cross-layer consistency of their underlying redundancy structures. Consequently, directly reusing historical clustering results may fail to capture changes in value similarities and cluster assignments. Since these changes are largely localized near cluster boundaries, the inherited clustering results can be efficiently refined using current-layer value representations.

We initialize the current-layer assignments as
$a_j^{(\ell,0)}=a_j^{(\ell-1)}$ and refine them by alternating
cluster center updates and local token reassignment:
\begin{equation}
\begin{aligned}
\mathbf c_k^{(\ell,t)}
&= \frac{1}{|C_k^{(\ell,t)}|}
\sum_{j:a_j^{(\ell,t)}=k}\mathbf v_j^{(\ell)},\\
a_j^{(\ell,t+1)}
&= \arg\max_{k\in\mathcal N_{a_j^{(\ell,t)}}^+}
\operatorname{cos}\left(
\mathbf v_j^{(\ell)},\mathbf c_k^{(\ell,t)}
\right),
\end{aligned}
\label{eq:local_cluster_refinement}
\end{equation}
where $C_k^{(\ell,t)}=\{j:a_j^{(\ell,t)}=k\}$
and $\mathcal N_k^+=\{k\}\cup\mathcal N_k$ includes
cluster $k$ and its neighbors. After refinement, the
representative
$r_C=\arg\max_{j\in C}\operatorname{cos}(
\mathbf v_j^{(\ell)},\mathbf c_C^{(\ell)})$
is selected, where $\mathbf c_C^{(\ell)}$ is the final cluster
center and $\mathbf v_{r_C}^{(\ell)}$ is the representative value.
To avoid repeated neighborhood construction, neighboring
clusters are determined only at the initial layer and reused
across subsequent layers.

\subsection{Joint Probability--Value Sparse Attention}
\label{sec:joint_sparse_attention}

With the estimated dense attention probabilities $\hat{p}_{ij}^{(\ell)}$ and refined value clusters $\mathcal{P}^{(\ell)}=\{C_1,\ldots,C_K\}$, we construct sparse attention masks by jointly considering entry importance and value redundancy. While attention probabilities indicate the importance of individual entries, value redundancy determines whether their contributions can be approximated by shared representations. Accordingly, we adopt value clusters as the basic units for sparsification and introduce four representation states: \textbf{Drop}, \textbf{Representative (Rep)}, \textbf{History}, and \textbf{Exact}. Drop discards all entries in a cluster, while Rep retains only its representative to approximate the remaining members. History preserves previously evaluated entries together with the representative, whereas Exact retains all entries without approximation. These states provide different trade-offs between representation fidelity and computational cost.

To evaluate these states, we define a probability-weighted value distortion that accounts for both attention importance and approximation error. For query $i$ and cluster $C$, the distortion under state $s$ is
\begin{equation}
\mathcal{L}_{iC}(s)=
\begin{cases}
\displaystyle
\sum_{j\in C}\hat{p}_{ij}^{(\ell)}
\|\mathbf{v}_j^{(\ell)}\|_2^2,
& s=\mathrm{Drop},\\[6pt]
\displaystyle
\sum_{j\in C}\hat{p}_{ij}^{(\ell)}d_{jC}^{(\ell)},
& s=\mathrm{Rep},\\[6pt]
\displaystyle
\sum_{j\in C\setminus\mathcal{H}_{iC}}
\hat{p}_{ij}^{(\ell)}d_{jC}^{(\ell)},
& s=\mathrm{History},\\[6pt]
0,
& s=\mathrm{Exact},
\end{cases}
\label{eq:cluster_state_loss}
\end{equation}
where $d_{jC}^{(\ell)}=\|\mathbf{v}_j^{(\ell)}-\mathbf{v}_{r_C}^{(\ell)}\|_2^2$ measures the value distortion relative to the cluster representative $r_C$, and $\mathcal{H}_{iC}=C\cap\mathcal{H}_i^{(\ell-1)}$ denotes the historical entries within cluster $C$. Since the value-dependent terms are query-independent, they can be computed once per layer and reused across queries. Moreover, the History loss can be obtained by subtracting the contributions of historical entries from the Rep loss, avoiding redundant computations.

\paragraph{Budget-Constrained State Optimization.}
Since value clusters exhibit varying attention importance and degrees of redundancy, assigning a uniform representation state may introduce unnecessary computations or excessive approximation errors. We therefore jointly optimize cluster representation states to minimize the total probability-weighted distortion under a fixed attention budget:
\begin{equation}
\begin{aligned}
\min_{\{s_{iC}\}}\quad
&\sum_C \mathcal{L}_{iC}(s_{iC}),\\
\mathrm{s.t.}\quad
&\sum_C c_{iC}(s_{iC})\leq B_i,
\end{aligned}
\label{eq:joint_state_optimization}
\end{equation}
where $s_{iC}\in\mathcal{S}_C$ denotes the selected state for cluster $C$, and $B_i$ specifies the total number of entries permitted for query $i$. The state cost $c_{iC}(s)$ is defined by the number of retained entries, given by $0$, $1$, $|\mathcal{H}_{iC}\cup\{r_C\}|$, and $|C|$ for Drop, Rep, History, and Exact, respectively. This formulation adaptively allocates computational resources across clusters, favoring inexpensive approximations for low-contribution clusters while preserving more entries where approximation would incur greater distortion.

Inspired by the penalized distortion formulation of DP-means~\cite{kulis2011revisiting}, we introduce a shared resource price $\lambda$ to balance representation distortion against the number of retained entries. For a fixed $\lambda$, the state selection decomposes across clusters as
\begin{equation}
s_{iC}^{*}(\lambda)
=
\arg\min_{s\in\mathcal{S}_C}
\left[
\mathcal{L}_{iC}(s)+\lambda c_{iC}(s)
\right].
\label{eq:penalized_state_selection}
\end{equation}
We determine $\lambda$ by bisection search to obtain a budget-feasible state allocation. To reduce overhead, neighboring queries are grouped into query groups that share the same retained-key set, amortizing state selection and mask construction costs. The selected states are converted into sparse attention masks. Finally, the resulting attention information and updated value clusters are propagated to subsequent layers as historical priors.

%% file: sec/4_experiements.tex
\input{table/ScanNetPose.tex}
\input{table/ScanNet3DRecon.tex}

\input{table/Sparse_Camera.tex}
\section{Experiments}
\subsection{Implementation Details}
% Experiments are conducted on three representative visual geometry foundation models, including VGGT~\cite{wang2025vggt}, $\pi^3$~\cite{wang2026pi}, and VGGT-$\Omega$~\cite{wang2026vggt}, with camera pose estimation and point-cloud reconstruction evaluated on ScanNet, 7Scenes, ETH3D, NRGBD, and DTU. Following FastVGGT~\cite{shen2026fastvggt} and SAF3R~\cite{deng2026saf3r}, 50 sequences are selected from the ScanNet to construct densely sampled inputs of 100, 300, 500, and 1000 frames, while the remaining datasets employ sparsely sampled inputs of up to 128 frames. Camera pose estimation is evaluated using AUC, and point-cloud reconstruction using Accuracy (Acc), Completeness (Comp), and Chamfer Distance (CD), as applicable. Inference efficiency is measured by inference latency (s), acceleration ratio (Spd.) relative to the corresponding dense backbone, and allocated GPU memory (Mem.). DuCo is implemented as a training-free module that sparsifies only global attention, without modifying pretrained model parameters. The overall entry sparsity across all global attention layers is set to 90\%, corresponding to a 10\% retention ratio. For probability estimation, each region contains 256 keys, with the consistency coefficient $\tau=0.5$. The number of value clusters is set to 10\% of the total patch token count. All experiments are conducted on a single NVIDIA H800 GPU.

Experiments use three visual geometry foundation models, VGGT~\cite{wang2025vggt}, $\pi^3$~\cite{wang2026pi}, and VGGT-$\Omega$~\cite{wang2026vggt}, with camera pose estimation and point-cloud reconstruction evaluated on ScanNet, 7Scenes, ETH3D, NRGBD, and DTU. Following FastVGGT~\cite{shen2026fastvggt} and SAF3R~\cite{deng2026saf3r}, 50 ScanNet test sequences form densely sampled 100-, 300-, 500-, and 1000-frame inputs, while other datasets use sparsely sampled inputs of up to 128 frames. AVGGT~\cite{sun2026avggt} is reproduced with a K/V subsampling factor of 4, while SAF3R uses the DINO Top-16 configuration. Camera pose estimation is evaluated using AUC, and point-cloud reconstruction using Accuracy (Acc), Completeness (Comp), and Chamfer Distance (CD), as applicable. Inference efficiency is measured by inference latency (s), acceleration ratio (Spd.) relative to the dense backbone, and allocated GPU memory (Mem.). DuCo is implemented as a training-free module that sparsifies only global attention, without modifying model parameters. The entry sparsity across all global attention layers is set to 90\%, corresponding to a 10\% retention ratio. For probability estimation, regions contain 256 keys, with consistency coefficient $\tau=0.5$. The number of value clusters is set to 10\% of the patch token count. All experiments are conducted on a single NVIDIA H800 GPU.

\subsection{Camera Pose Estimation}

Table~\ref{tab:scannet_pose} presents camera pose estimation results on densely sampled ScanNet50 sequences with 100 to 1000 frames. DuCo achieves increasing acceleration with sequence length, reaching 5.26$\times$, 5.51$\times$, and 4.47$\times$ speedups on VGGT, $\pi^3$, and VGGT-$\Omega$, respectively, at 1000 frames. On VGGT, AUC@30 remains nearly unchanged across all sequence lengths, with 74.12 at 1000 frames compared to 74.08 for dense inference. On $\pi^3$, DuCo consistently achieves the highest AUC@3 and AUC@30 among accelerated methods, attaining 25.92/86.95 at 1000 frames. On VGGT-$\Omega$, the replacement of some global attention layers with register-only layers reduces the available acceleration opportunities and introduces larger gaps between successive global attention layers, potentially weakening the reliability of historical priors. Nevertheless, DuCo achieves a 4.47$\times$ speedup while maintaining comparable pose accuracy (10.00/77.69 versus 10.10/77.63 for dense inference). Moreover, DuCo completes all 1000-frame settings, whereas SparseVGGT and SAF3R encounter OOM errors.

Table~\ref{tab:pose_reconstruction} further evaluates camera pose estimation under sparsely sampled views on 7Scenes, ETH3D, NRGBD, and DTU. On 7Scenes, DuCo improves AUC@3 from 20.32 to 20.68 relative to dense VGGT, while maintaining comparable AUC@30 (83.82 versus 83.89). On ETH3D, it achieves 21.63/77.33, remaining competitive with FastVGGT and outperforming AVGGT, SparseVGGT, and SAF3R. On NRGBD, DuCo obtains 52.19/92.89, outperforming SparseVGGT and SAF3R, although an accuracy gap remains relative to dense VGGT. On DTU, it achieves 67.61/96.26, maintaining pose accuracy close to the dense baseline of 69.21/96.46. These results demonstrate that DuCo maintains competitive pose estimation accuracy across both dense and sparse multi-view settings, with dataset-dependent accuracy trade-offs.

\subsection{Point-Cloud Reconstruction}

Table~\ref{tab:scannet_pointmap} reports point-cloud reconstruction results on densely sampled ScanNet50 sequences. On VGGT, DuCo achieves lower Chamfer Distance (CD) than dense inference across all sequence lengths, reducing CD from 0.284 to 0.279 at 1000 frames with a 5.26$\times$ speedup. Similarly, on $\pi^3$, CD remains comparable to or better than the dense baseline, reaching 0.396 versus 0.398 at 1000 frames with 5.51$\times$ acceleration. On VGGT-$\Omega$, CD increases from 0.367 to 0.384 at 1000 frames, alongside a 4.47$\times$ speedup. Overall, DuCo substantially reduces global attention computation while largely preserving reconstruction quality, with architecture-dependent accuracy–efficiency trade-offs.

Table~\ref{tab:pose_reconstruction} further evaluates reconstruction quality under sparsely sampled views across four benchmark datasets. 
On 7Scenes, DuCo improves reconstruction accuracy from 0.156 to 0.142 relative to dense VGGT, with a slight increase in completeness error from 0.124 to 0.131. On ETH3D, it achieves the lowest completeness error of 0.879 among accelerated methods, outperforming FastVGGT, SparseVGGT, AVGGT, and SAF3R. On NRGBD, DuCo obtains Acc/Comp of 0.019/0.013, matching FastVGGT in accuracy and dense VGGT in completeness. On DTU, it preserves the dense backbone's reconstruction performance with Acc/Comp of 0.003/0.001. Overall, these results indicate that jointly considering attention probabilities and value redundancy enables DuCo to retain useful geometric information under attention sparsification.

\subsection{Ablation Study}
\input{table/ablation_study.tex}
Table~\ref{tab:core_ablation} evaluates the contributions of history-guided probability estimation, value-aware state selection, and local cluster refinement. Under the same nominal budgets, replacing regional with historical probability estimation improves AUC@3 from 11.18 to 17.44 and reduces CD from 0.158 to 0.152, demonstrating the benefit of historical information for entry selection. By introducing value clustering and Drop/Rep/Exact states, variant (c) further improves AUC@3 to 19.76 and CD to 0.139 relative to variant (a), highlighting the importance of value-aware approximation. The full DuCo additionally incorporates historical probabilities and the History state, achieving 20.68 AUC@3 and 0.137 CD. Compared with global clustering in variant (d), local refinement improves AUC@3 from 19.88 to 20.68 while reducing latency from 3.83 to 2.69 s. Overall, DuCo surpasses dense inference in both geometric metrics while substantially reducing computational cost.

\begin{figure}[htbp]
    \centering
    \begin{subfigure}{\columnwidth}
        \centering
        \includegraphics[width=\linewidth]{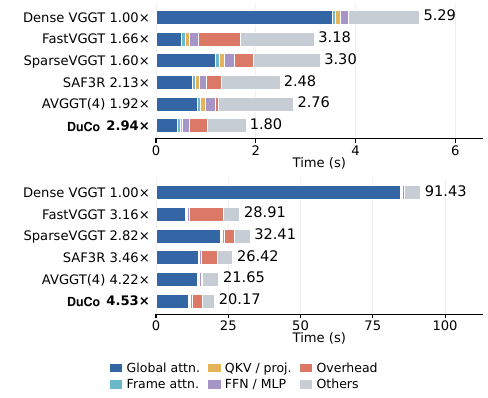}
        \caption{Full-forward breakdown.}
        \label{fig:time_breakdown_full}
    \end{subfigure}
    \par\vspace{3pt}
    \begin{subfigure}{\columnwidth}
        \centering
        \includegraphics[width=\linewidth]{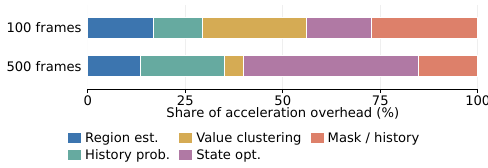}
        \caption{Composition of our acceleration overhead.}
        \label{fig:ours_overhead_breakdown}
    \end{subfigure}
    \caption{Runtime breakdown on ScanNet50. (a) End-to-end inference latency at 100 and 500 frames. (b) Composition of DuCo's sparsification overhead. Total latency is averaged over 50 scenes, with component proportions profiled on 10 sequences.}
    \vspace{-6mm}
    \label{fig:time_breakdown}
\end{figure}

\subsection{Efficiency Analysis}

Figure~\ref{fig:time_breakdown_full} compares the inference time breakdown across different methods. As the sequence length increases, global attention becomes the dominant computational bottleneck. DuCo achieves 2.94$\times$ and 4.53$\times$ speedups over dense VGGT with 100 and 500 frames, respectively. Unlike layer-independent sparsification methods, DuCo reduces overhead by reusing previously computed information. Specifically, historical probability estimates are derived during preceding sparse attention computations, while local clustering and query grouping reduce redundant operations. Moreover, although attention probabilities vary across queries, the value-distortion terms are query-independent and can be precomputed and shared across queries and cluster representation states, avoiding repeated value comparisons. These designs keep overhead limited relative to the savings in attention computation. Nevertheless, as shown in Figure~\ref{fig:ours_overhead_breakdown}, the proportion of overhead attributed to state optimization increases with sequence length, reflecting the growing number of cluster-state decisions.

%% file: table/ScanNetPose.tex
\begin{table*}[!t]
\centering
\setlength{\tabcolsep}{0pt}
\renewcommand{\arraystretch}{1.15}
\fontsize{8}{9.6}\selectfont
\caption{ScanNet comparison using AUC@3, AUC@30, acceleration ratio (Spd.), and allocated GPU memory (Mem.) under 100-, 300-, 500-, and 1000-frame input settings. Spd. is measured relative to the corresponding backbone baseline; Mem. is reported in GiB, where lower is better. Timings use native shared execution and same-GPU pairing. SAF3R uses DINO per-frame Top-16.}
\label{tab:scannet_pose}
\vspace{4pt}
\begin{tabular*}{\textwidth}{@{\extracolsep{\fill}}l*{16}{c}@{}}
\toprule
\multirow{2}{*}{Method}
& \multicolumn{4}{c}{100 Frames}
& \multicolumn{4}{c}{300 Frames}
& \multicolumn{4}{c}{500 Frames}
& \multicolumn{4}{c}{1000 Frames} \\
\cmidrule(lr){2-5}\cmidrule(lr){6-9}\cmidrule(lr){10-13}\cmidrule(lr){14-17}
& A$_3\uparrow$ & A$_{30}\uparrow$ & Spd.$\uparrow$ & Mem.$\downarrow$
& A$_3\uparrow$ & A$_{30}\uparrow$ & Spd.$\uparrow$ & Mem.$\downarrow$
& A$_3\uparrow$ & A$_{30}\uparrow$ & Spd.$\uparrow$ & Mem.$\downarrow$
& A$_3\uparrow$ & A$_{30}\uparrow$ & Spd.$\uparrow$ & Mem.$\downarrow$ \\
\midrule

\rowcolor{gray!10}[3pt][3pt]
\makecell[l]{VGGT~\cite{wang2025vggt}\\[-2pt]{\scriptsize\color{black!45} CVPR'25}}
& 13.88 & 79.47 & 1.00 & 6.50
& 13.91 & 78.55 & 1.00 & 15.12
& 13.37 & 77.02 & 1.00 & 23.75
& 10.78 & 74.08 & 1.00 & 45.29 \\

\makecell[l]{FastVGGT~\cite{shen2026fastvggt}\\[-2pt]{\scriptsize\color{black!45} ICLR'26}}
& \textbf{14.27} & \underline{79.18} & 1.66 & 8.34
& \underline{13.67} & \textbf{78.91} & 2.74 & 20.64
& \textbf{13.84} & \textbf{78.22} & 3.16 & 32.94
& \textbf{12.90} & \textbf{78.34} & 4.27 & 63.67 \\

\makecell[l]{AVGGT~\cite{sun2026avggt}\\[-2pt]{\scriptsize\color{black!45} CVPR'26}}
& 13.22 & 78.44 & 1.92 & \textbf{6.50}
& 12.99 & 78.23 & \underline{3.49} & \textbf{15.12}
& 13.12 & \underline{77.33} & \underline{4.22} & \textbf{23.75}
& \underline{12.17} & \underline{77.12} & \underline{5.02} & \textbf{45.29} \\

\makecell[l]{SparseVGGT~\cite{wang2026block}\\[-2pt]{\scriptsize\color{black!45} CVPR'26}}
& 6.94 & 74.93 & 1.60 & 12.42
& 6.65 & 73.01 & 2.34 & 26.75
& 6.35 & 69.39 & 2.82 & 42.33
& - & - & - & OOM \\

\makecell[l]{SAF3R~\cite{deng2026saf3r}\\[-2pt]{\scriptsize\color{black!45} ECCV'26}}
& 12.86 & 78.20 & \underline{2.19} & 9.22
& 12.31 & 76.46 & 3.09 & 26.98
& 12.24 & 75.91 & 3.46 & 49.67
& - & - & - & OOM \\

\rowcolor{gray!10}[3pt][3pt]
\textbf{DuCo}
& \underline{14.16} & \textbf{79.35} & \textbf{2.94} & \underline{6.97}
& \textbf{14.09} & \underline{78.61} & \textbf{4.05} & \underline{16.59}
& \underline{13.21} & 77.16 & \textbf{4.53} & \underline{26.90}
& 10.94 & 74.12 & \textbf{5.26} & \underline{53.74} \\

\midrule

\rowcolor{gray!10}[3pt][3pt]
\makecell[l]{$\pi^3$~\cite{wang2026pi}\\[-2pt]{\scriptsize\color{black!45} ICLR'26}}
& 29.08 & 88.00 & 1.00 & 9.40
& 28.83 & 87.61 & 1.00 & 17.46
& 29.22 & 87.15 & 1.00 & 25.52
& 28.82 & 87.71 & 1.00 & 45.67 \\

\makecell[l]{FastVGGT~\cite{shen2026fastvggt}\\[-2pt]{\scriptsize\color{black!45} ICLR'26}}
& 23.23 & 86.17 & 1.92 & \textbf{9.40}
& 24.85 & \underline{86.46} & 2.82 & \underline{17.82}
& 24.51 & 85.78 & 3.18 & \underline{26.37}
& \underline{25.12} & \underline{86.64} & 4.31 & \underline{47.75} \\

\makecell[l]{AVGGT~\cite{sun2026avggt}\\[-2pt]{\scriptsize\color{black!45} CVPR'26}}
& 22.67 & 86.20 & \underline{2.36} & \textbf{9.40}
& 22.42 & 85.97 & \underline{4.01} & \textbf{17.46}
& 21.81 & 84.98 & \textbf{4.86} & \textbf{25.52}
& 22.37 & 85.83 & \textbf{6.00} & \textbf{45.67} \\

\makecell[l]{SparseVGGT~\cite{wang2026block}\\[-2pt]{\scriptsize\color{black!45} CVPR'26}}
& 7.15 & 75.93 & 1.96 & 9.60
& 6.78 & 75.69 & 2.68 & 22.23
& 6.97 & 72.65 & 2.99 & 32.93
& - & - & - & OOM \\

\makecell[l]{SAF3R~\cite{deng2026saf3r}\\[-2pt]{\scriptsize\color{black!45} ECCV'26}}
& \underline{25.04} & \underline{86.66} & 2.02 & 10.41
& \underline{24.99} & 86.11 & 3.12 & 25.73
& \underline{25.91} & \underline{85.93} & 3.57 & 52.45
& - & - & - & OOM \\

\rowcolor{gray!10}[3pt][3pt]
\textbf{DuCo}
& \textbf{25.25} & \textbf{86.98} & \textbf{2.48} & \underline{9.59}
& \textbf{25.81} & \textbf{86.68} & \textbf{4.02} & 18.06
& \textbf{26.05} & \textbf{86.23} & \underline{4.68} & 26.76
& \textbf{25.92} & \textbf{86.95} & \underline{5.51} & 48.96 \\

\midrule

\rowcolor{gray!10}[3pt][3pt]
\makecell[l]{VGGT-$\Omega$~\cite{wang2026vggt}\\[-2pt]{\scriptsize\color{black!45} CVPR'26}}
& 11.27 & 79.20 & 1.00 & 11.62
& 10.98 & 78.27 & 1.00 & 22.94
& 10.67 & 77.00 & 1.00 & 34.26
& 10.10 & 77.63 & 1.00 & 62.58 \\

\makecell[l]{FastVGGT~\cite{shen2026fastvggt}\\[-2pt]{\scriptsize\color{black!45} ICLR'26}}
& \underline{9.85} & \underline{77.98} & 1.64 & 12.17
& \underline{10.24} & \textbf{77.82} & 2.42 & 24.55
& \underline{10.22} & \textbf{77.29} & 2.79 & 36.94
& \underline{9.64} & \textbf{77.82} & 3.74 & 67.88 \\

\makecell[l]{AVGGT~\cite{sun2026avggt}\\[-2pt]{\scriptsize\color{black!45} CVPR'26}}
& 9.28 & 77.42 & \textbf{1.89} & \underline{11.62}
& 9.41 & 77.26 & \textbf{3.09} & \underline{22.94}
& 9.35 & 76.73 & \textbf{3.74} & \underline{34.27}
& 8.62 & 77.27 & \textbf{4.53} & \underline{62.57} \\

\makecell[l]{SparseVGGT~\cite{wang2026block}\\[-2pt]{\scriptsize\color{black!45} CVPR'26}}
& 7.88 & 76.23 & \underline{1.68} & 12.05
& 7.65 & 74.39 & 2.37 & 24.76
& 7.71 & 72.78 & 2.63 & 38.63
& - & - & - & OOM \\

\rowcolor{gray!10}[3pt][3pt]
\textbf{DuCo}
& \textbf{10.71} & \textbf{78.55} & 1.41 & \textbf{6.01}
& \textbf{10.78} & \underline{77.72} & \underline{2.45} & \textbf{13.64}
& \textbf{10.77} & \underline{77.08} & \underline{3.22} & \textbf{21.81}
& \textbf{10.00} & \underline{77.69} & \underline{4.47} & \textbf{42.33} \\

\bottomrule
\end{tabular*}
\vspace{-10pt}
\end{table*}

%% file: table/ScanNet3DRecon.tex
\begin{table*}[!t]
\centering
\setlength{\tabcolsep}{1pt}
\renewcommand{\arraystretch}{1.15}
\fontsize{8}{9.6}\selectfont
\caption{Point-cloud reconstruction results on ScanNet with 100, 300, 500, and 1000 input frames. Acc, Comp, and CD denote mean accuracy, completeness, and Chamfer distance, respectively. Spd. is the acceleration ratio relative to the corresponding backbone baseline. Timings use native shared execution and same-GPU pairing. SAF3R uses DINO per-frame Top-16.}
\label{tab:scannet_pointmap}
\vspace{4pt}
\begin{tabular*}{\textwidth}{@{\extracolsep{\fill}}l*{16}{c}@{}}
\toprule
\multirow{2}{*}{Method}
& \multicolumn{4}{c}{100 Frames}
& \multicolumn{4}{c}{300 Frames}
& \multicolumn{4}{c}{500 Frames}
& \multicolumn{4}{c}{1000 Frames} \\
\cmidrule(lr){2-5}\cmidrule(lr){6-9}\cmidrule(lr){10-13}\cmidrule(lr){14-17}
& Acc$\downarrow$ & Comp$\downarrow$ & CD$\downarrow$ & Spd.$\uparrow$
& Acc$\downarrow$ & Comp$\downarrow$ & CD$\downarrow$ & Spd.$\uparrow$
& Acc$\downarrow$ & Comp$\downarrow$ & CD$\downarrow$ & Spd.$\uparrow$
& Acc$\downarrow$ & Comp$\downarrow$ & CD$\downarrow$ & Spd.$\uparrow$ \\
\midrule

% ==================== VGGT ====================
\rowcolor{gray!10}[3pt][3pt]
\makecell[l]{VGGT~\cite{wang2025vggt}\\[-2pt]{\scriptsize\color{black!45} CVPR'25}}
& 0.226 & 0.301 & 0.263 & 1.00
& 0.236 & 0.314 & 0.275 & 1.00
& 0.234 & 0.322 & 0.278 & 1.00
& 0.237 & 0.330 & 0.284 & 1.00 \\
\makecell[l]{FastVGGT~\cite{shen2026fastvggt}\\[-2pt]{\scriptsize\color{black!45} ICLR'26}}
& \underline{0.223} & \underline{0.290} & \underline{0.256} & 1.66
& \underline{0.233} & \underline{0.308} & \underline{0.271} & 2.74
& 0.233 & 0.318 & 0.275 & 3.16
& \underline{0.236} & 0.334 & 0.285 & 4.27 \\
% DA-VGGT
% & 0.021 & \textbf{0.019} & 0.020 & 0.86
% & 0.025 & \textbf{0.019} & 0.022 & 2.04
% & \underline{0.030} & \textbf{0.022} & \textbf{0.026} & 2.74
% & 0.035 & \textbf{0.023} & \textbf{0.029} & 2.71 \\
\makecell[l]{AVGGT~\cite{sun2026avggt}\\[-2pt]{\scriptsize\color{black!45} CVPR'26}}
& \underline{0.223} & \textbf{0.284} & \textbf{0.254} & 1.92
& \textbf{0.230} & \textbf{0.299} & \textbf{0.265} & \underline{3.49}
& \textbf{0.227} & \textbf{0.302} & \textbf{0.265} & \underline{4.22}
& \textbf{0.233} & \textbf{0.320} & \textbf{0.276} & \underline{5.02} \\
\makecell[l]{SparseVGGT~\cite{wang2026block}\\[-2pt]{\scriptsize\color{black!45} CVPR'26}}
& 0.230 & 0.301 & 0.266 & 1.60
& 0.241 & 0.316 & 0.278 & 2.34
& 0.238 & \underline{0.315} & 0.277 & 2.82
& - & - & - & OOM \\
\makecell[l]{SAF3R~\cite{deng2026saf3r}\\[-2pt]{\scriptsize\color{black!45} ECCV'26}}
& \textbf{0.221} & \underline{0.290} & \underline{0.256} & \underline{2.19}
& 0.239 & 0.313 & 0.276 & 3.09
& 0.236 & 0.320 & 0.278 & 3.46
& - & - & - & OOM \\
\rowcolor{gray!10}[3pt][3pt]
\textbf{DuCo}
& \underline{0.223} & 0.299 & 0.261 & \textbf{2.94}
& \underline{0.233} & 0.313 & 0.273 & \textbf{4.05}
& \underline{0.229} & \underline{0.315} & \underline{0.272} & \textbf{4.53}
& \textbf{0.233} & \underline{0.325} & \underline{0.279} & \textbf{5.26} \\
\midrule

% ==================== pi^3 ====================
\rowcolor{gray!10}[3pt][3pt]
\makecell[l]{$\pi^3$~\cite{wang2026pi}\\[-2pt]{\scriptsize\color{black!45} ICLR'26}}
& 0.352 & 0.458 & 0.405 & 1.00
& 0.350 & 0.471 & 0.411 & 1.00
& 0.342 & 0.451 & 0.397 & 1.00
& 0.342 & 0.455 & 0.398 & 1.00 \\
\makecell[l]{FastVGGT~\cite{shen2026fastvggt}\\[-2pt]{\scriptsize\color{black!45} ICLR'26}}
& 0.350 & \underline{0.448} & 0.399 & 1.92
& 0.346 & 0.475 & 0.411 & 2.82
& 0.338 & 0.451 & \underline{0.394} & 3.18
& \underline{0.338} & 0.457 & 0.398 & 4.31 \\
% DA-VGGT
% & 0.017 & \textbf{0.018} & \textbf{0.017} & 0.74
% & 0.020 & \textbf{0.020} & \textbf{0.020} & 1.36
% & 0.022 & \textbf{0.021} & \textbf{0.021} & 1.71
% & 0.030 & \textbf{0.022} & 0.026 & 1.99 \\
\makecell[l]{AVGGT~\cite{sun2026avggt}\\[-2pt]{\scriptsize\color{black!45} CVPR'26}}
& \textbf{0.341} & \textbf{0.434} & \textbf{0.387} & \underline{2.36}
& \underline{0.342} & \underline{0.467} & 0.405 & \underline{4.01}
& \underline{0.336} & \textbf{0.439} & \textbf{0.387} & \textbf{4.86}
& \textbf{0.333} & \textbf{0.442} & \textbf{0.387} & \textbf{6.00} \\
\makecell[l]{SparseVGGT~\cite{wang2026block}\\[-2pt]{\scriptsize\color{black!45} CVPR'26}}
& \underline{0.344} & 0.451 & \underline{0.397} & 1.96
& \textbf{0.333} & \textbf{0.461} & \textbf{0.397} & 2.68
& \textbf{0.325} & \underline{0.448} & \textbf{0.387} & 2.99
& - & - & - & OOM \\
\makecell[l]{SAF3R~\cite{deng2026saf3r}\\[-2pt]{\scriptsize\color{black!45} ECCV'26}}
& 0.361 & 0.461 & 0.411 & 2.02
& \underline{0.342} & 0.468 & 0.405 & 3.12
& 0.348 & 0.460 & 0.404 & 3.57
& - & - & - & OOM \\
\rowcolor{gray!10}[3pt][3pt]
\textbf{DuCo}
& 0.357 & 0.454 & 0.405 & \textbf{2.48}
& 0.346 & \textbf{0.461} & \underline{0.404} & \textbf{4.02}
& \underline{0.336} & 0.452 & \underline{0.394} & \underline{4.68}
& 0.341 & \underline{0.451} & \underline{0.396} & \underline{5.51} \\
\midrule

% ==================== VGGT-Omega ====================
\rowcolor{gray!10}[3pt][3pt]
\makecell[l]{VGGT-$\Omega$~\cite{wang2026vggt}\\[-2pt]{\scriptsize\color{black!45} CVPR'26}}
& 0.262 & 0.430 & 0.346 & 1.00
& 0.265 & 0.455 & 0.360 & 1.00
& 0.267 & 0.465 & 0.366 & 1.00
& 0.264 & 0.470 & 0.367 & 1.00 \\
\makecell[l]{FastVGGT~\cite{shen2026fastvggt}\\[-2pt]{\scriptsize\color{black!45} ICLR'26}}
& \textbf{0.258} & \textbf{0.405} & \textbf{0.332} & 1.64
& \underline{0.265} & \underline{0.444} & \underline{0.354} & 2.42
& \textbf{0.264} & \textbf{0.452} & \textbf{0.358} & 2.79
& \textbf{0.263} & \textbf{0.461} & \textbf{0.362} & 3.74 \\
% DA-VGGT
% & 0.023 & \textbf{0.019} & 0.021 & 0.59
% & 0.029 & \textbf{0.019} & 0.024 & 0.96
% & 0.031 & \textbf{0.023} & 0.029 & 1.12
% & 0.039 & \textbf{0.025} & 0.032 & 1.50 \\
\makecell[l]{AVGGT~\cite{sun2026avggt}\\[-2pt]{\scriptsize\color{black!45} CVPR'26}}
& \underline{0.263} & \underline{0.415} & \underline{0.339} & \textbf{1.89}
& \textbf{0.264} & \textbf{0.442} & \textbf{0.353} & \textbf{3.09}
& \underline{0.267} & \underline{0.460} & \underline{0.363} & \textbf{3.74}
& \underline{0.264} & \underline{0.464} & \underline{0.364} & \textbf{4.53} \\
\makecell[l]{SparseVGGT~\cite{wang2026block}\\[-2pt]{\scriptsize\color{black!45} CVPR'26}}
& \underline{0.263} & 0.423 & 0.343 & \underline{1.68}
& 0.274 & 0.458 & 0.366 & 2.37
& 0.284 & 0.491 & 0.387 & 2.63
& - & - & - & OOM \\
\rowcolor{gray!10}[3pt][3pt]
\textbf{DuCo}
& \underline{0.263} & 0.441 & 0.352 & 1.41
& 0.271 & 0.464 & 0.367 & \underline{2.45}
& 0.277 & 0.488 & 0.382 & \underline{3.22}
& 0.276 & 0.491 & 0.384 & \underline{4.47} \\
\bottomrule
\end{tabular*}
\vspace{-5pt}
\end{table*}

%% file: table/Sparse_Camera.tex
\begin{table*}[!t]
\centering
\setlength{\tabcolsep}{0pt}
\renewcommand{\arraystretch}{1.15}
\fontsize{8}{9.6}\selectfont
\caption{\textbf{Camera pose estimation and 3D reconstruction on four benchmarks under sparse views.} $A_3$ and $A_{30}$ are pose AUC at thresholds 3 and 30; Acc. and Comp. are reconstruction accuracy and completeness. Higher AUC and lower errors are better. Bold and underline mark the best and second-best accelerated results.}
\label{tab:pose_reconstruction}
\begin{tabular*}{\textwidth}{@{\extracolsep{\fill}}l*{16}{c}@{}}
\toprule
\multirow{2}{*}{\textbf{Methods}}
& \multicolumn{4}{c}{\textbf{7Scenes}}
& \multicolumn{4}{c}{\textbf{ETH3D}}
& \multicolumn{4}{c}{\textbf{NRGBD}}
& \multicolumn{4}{c}{\textbf{DTU}} \\
\cmidrule(lr){2-5}
\cmidrule(lr){6-9}
\cmidrule(lr){10-13}
\cmidrule(lr){14-17}
& A$_3\uparrow$ & A$_{30}\uparrow$ & Acc.$\downarrow$ & Comp.$\downarrow$
& A$_3\uparrow$ & A$_{30}\uparrow$ & Acc.$\downarrow$ & Comp.$\downarrow$
& A$_3\uparrow$ & A$_{30}\uparrow$ & Acc.$\downarrow$ & Comp.$\downarrow$
& A$_3\uparrow$ & A$_{30}\uparrow$ & Acc.$\downarrow$ & Comp.$\downarrow$ \\
\midrule

\rowcolor{gray!10}
VGGT~\cite{wang2025vggt}
& 20.32 & 83.89 & 0.156 & 0.124
& 24.85 & 80.10 & 0.317 & 0.831
& 58.20 & 94.34 & 0.016 & 0.013
& 69.21 & 96.46 & 0.003 & 0.001 \\
FastVGGT~\cite{shen2026fastvggt}
& 18.81 & 82.98 & \underline{0.125} & 0.137
& \textbf{21.95} & \textbf{77.47} & \textbf{0.367} & \underline{0.911}
& \underline{56.52} & \underline{93.74} & \textbf{0.019} & \underline{0.014}
& 67.23 & 96.25 & \underline{0.004} & \textbf{0.001} \\
SparseVGGT~\cite{wang2026block}
& 14.40 & 80.96 & 0.157 & 0.165
& 11.06 & 64.08 & 0.443 & 1.764
& 33.70 & 90.62 & 0.036 & 0.022
& 58.30 & 95.71 & 0.005 & \underline{0.002} \\
AVGGT~\cite{sun2026avggt}
& 19.03 & 83.36 & \textbf{0.120} & 0.143
& 19.93 & 73.04 & 0.444 & 1.061
& \textbf{56.73} & \textbf{93.96} & \underline{0.021} & 0.016
& 63.18 & 95.74 & 0.005 & \underline{0.002} \\
SAF3R~\cite{deng2026saf3r}
& \underline{20.56} & \textbf{83.92} & 0.147 & \textbf{0.129}
& 17.29 & 67.46 & 0.379 & 1.151
& 48.42 & 86.99 & 0.050 & 0.047
& \textbf{69.69} & \textbf{96.55} & \textbf{0.003} & \textbf{0.001} \\
\rowcolor{gray!12}
\textbf{DuCo}
& \textbf{20.68} & \underline{83.82} & 0.142 & \underline{0.131}
& \underline{21.63} & \underline{77.33} & \underline{0.368} & \textbf{0.879}
& 52.19 & 92.89 & \textbf{0.019} & \textbf{0.013}
& \underline{67.61} & \underline{96.26} & \textbf{0.003} & \textbf{0.001} \\

% \midrule

% \rowcolor{gray!12}
% $\pi^3$
% & 24.09 & 86.15 & -- & --
% & 32.70 & 85.57 & -- & --
% & -- & -- & -- & --
% & 62.95 & 94.85 & -- & -- \\
% Fast$\pi^3$
% & 23.80 & 85.30 & -- & --
% & 24.20 & 79.32 & -- & --
% & -- & -- & -- & --
% & 54.24 & 93.03 & -- & -- \\
% Sparse$\pi^3$
% & 12.25 & 81.15 & -- & --
% & 14.37 & 75.36 & -- & --
% & -- & -- & -- & --
% & 46.30 & 92.90 & -- & -- \\
% A$\pi^3$
% & \textbf{24.22} & \textbf{86.07} & -- & --
% & 22.74 & 78.03 & -- & --
% & -- & -- & -- & --
% & 55.30 & 93.45 & -- & -- \\
% SAF3R
% & 23.48 & 85.85 & -- & --
% & \textbf{24.30} & \textbf{83.03} & -- & --
% & -- & -- & -- & --
% & \textbf{61.46} & \textbf{94.79} & -- & -- \\

% \midrule

% \rowcolor{gray!12}
% DA3
% & 28.59 & 86.71 & -- & --
% & 48.37 & 91.40 & -- & --
% & -- & -- & -- & --
% & 93.99 & 99.38 & -- & -- \\
% FastDA3
% & 28.53 & 85.90 & -- & --
% & \textbf{37.89} & \textbf{88.53} & -- & --
% & -- & -- & -- & --
% & 88.69 & 98.82 & -- & -- \\
% ADA3
% & 25.05 & 85.15 & -- & --
% & 16.60 & 64.13 & -- & --
% & -- & -- & -- & --
% & 77.91 & 97.36 & -- & -- \\
% SAF3R
% & \textbf{28.84} & \textbf{86.80} & -- & --
% & 35.39 & 87.85 & -- & --
% & -- & -- & -- & --
% & \textbf{90.39} & \textbf{98.99} & -- & -- \\

\bottomrule
\end{tabular*}
\vspace{-2mm}
\end{table*}

%% file: table/ablation_study.tex
\begin{table}[t]
\centering
\setlength{\tabcolsep}{0pt}
\renewcommand{\arraystretch}{1.15}
\fontsize{8}{9.6}\selectfont
\caption{Core component ablation on 7Scenes (128 views).
Reg./Hist. denote regional/historical probability estimation;
D/R/H/E denote Drop/Representative/History/Exact states.
All sparse variants share the same nominal budgets.
Variant (c) disables historical key membership.}
\label{tab:core_ablation}
\vspace{4pt}
\begin{tabular*}{\columnwidth}{@{\extracolsep{\fill}}lcccrcr@{}}
\toprule
Variant & Prob. & States & Cluster
& AUC@3$\uparrow$ & CD$\downarrow$ & Time$\downarrow$ \\
\midrule
\rowcolor{gray!10}
Dense & -- & -- & --
& 20.32 & 0.140 & 7.97 \\
\midrule
(a) & Reg. & Top-$B$ & --
& 11.18 & 0.158 & \textbf{2.21} \\
(b) & Hist. & Top-$B$ & --
& 17.44 & 0.152 & 2.51 \\
(c) & Reg. & D/R/E & Local
& 19.76 & 0.139 & 2.64 \\
(d) & Hist. & D/R/H/E & Global
& 19.88 & 0.140 & 3.83 \\
\midrule
\rowcolor{gray!10}
\textbf{(e) Full} & Hist. & D/R/H/E & Local
& \textbf{20.68} & \textbf{0.137} & 2.69 \\
\bottomrule
\end{tabular*}
\vspace{-2mm}
\end{table}

%% file: sec/conclusion.tex
\section{Conclusion}

We present DuCo, a training-free sparse attention framework motivated by dual cross-layer consistency in global attention. Similarity in attention probability distributions and localized changes in value redundancy suggest that preceding-layer information can guide sparsification. DuCo estimates current-layer dense attention probabilities by compensating for unobserved entries in historical sparse computations and adapting historical importance to regional distribution shifts. Value clustering captures redundancy, while local refinement updates inherited clusters as features evolve. The estimated probabilities and refined clusters are coupled through probability-weighted value distortion, which guides the selection of four representation states under an attention budget. Experiments on three visual geometry foundation models and five benchmarks confirm that this treatment improves inference efficiency while maintaining competitive geometric accuracy.

%% file: main.bib
@String(CVPR= {IEEE Conf. Comput. Vis. Pattern Recog.})

@String(CVPR  = {CVPR})

@inproceedings{wang2025vggt,
  title={{VGGT}: Visual Geometry Grounded Transformer},
  author={Wang, Jianyuan and Chen, Minghao and Karaev, Nikita and Vedaldi, Andrea and Rupprecht, Christian and Novotny, David},
  booktitle={2025 IEEE/CVF Conference on Computer Vision and Pattern Recognition (CVPR)},
  pages={5294--5306},
  year={2025},
  organization={IEEE}
}

@inproceedings{shen2026fastvggt,
  title={{FastVGGT}: Fast Visual Geometry Transformer},
  author={Shen, You and Zhang, Zhipeng and Qu, Yansong and Zheng, Xiawu and Ji, Jiayi and Zhang, Shengchuan and Cao, Liujuan},
  booktitle={The Fourteenth International Conference on Learning Representations},
  year={2026}
}

@inproceedings{wang2026block,
  title={Block-sparse global attention for efficient multi-view geometry transformers},
  author={Wang, Chung-Shien Brian and Schmidt, Christian and Piekenbrinck, Jens and Leibe, Bastian},
  booktitle={Proceedings of the IEEE/CVF Conference on Computer Vision and Pattern Recognition},
  pages={14546--14555},
  year={2026}
}

@inproceedings{wang2026pi,
  title={{$\pi^3$}: Permutation-Equivariant Visual Geometry Learning},
  author={Wang, Yifan and Zhou, Jianjun and Zhu, Haoyi and Chang, Wenzheng and Zhou, Yang and Li, Zizun and Chen, Junyi and Pang, Jiangmiao and Shen, Chunhua and He, Tong},
  booktitle={International Conference on Learning Representations},
  volume={2026},
  pages={10481--10497},
  year={2026}
}

@article{wang2026vggt,
  title={{VGGT}-{$\Omega$}},
  author={Wang, Jianyuan and Chen, Minghao and Zhang, Shangzhan and Karaev, Nikita and Sch{\"o}nberger, Johannes and Labatut, Patrick and Bojanowski, Piotr and Novotny, David and Vedaldi, Andrea and Rupprecht, Christian},
  journal={arXiv preprint arXiv:2605.15195},
  year={2026}
}

@inproceedings{chen2026co,
  title={Co-Me: Confidence Guided Token Merging for Visual Geometric Transformers},
  author={Chen, Yutian and Qiu, Yuheng and Li, Ruogu and Patrikar, Jay and Scherer, Sebastian},
  booktitle={Proceedings of the IEEE/CVF Conference on Computer Vision and Pattern Recognition},
  pages={14590--14599},
  year={2026}
}

@inproceedings{sun2026avggt,
  title={{AVGGT}: Rethinking global attention for accelerating {VGGT}},
  author={Sun, Xianbing and Zhu, Zhikai and Lou, Zhengyu and Yang, Bo and Tang, Jinyang and Zhang, Liqing and Wang, He and Zhang, Jianfu},
  booktitle={Proceedings of the IEEE/CVF Conference on Computer Vision and Pattern Recognition},
  pages={251--260},
  year={2026}
}

@inproceedings{wang2026httm,
  title={Httm: Head-wise temporal token merging for faster vggt},
  author={Wang, Weitian and Meiner, Lukas and Shubham, Rai and De La Parra, Cecilia and Kumar, Akash},
  booktitle={Proceedings of the IEEE/CVF Conference on Computer Vision and Pattern Recognition},
  pages={26379--26388},
  year={2026}
}

@article{lin2025depth,
  title={Depth anything 3: Recovering the visual space from any views},
  author={Lin, Haotong and Chen, Sili and Liew, Junhao and Chen, Donny Y and Li, Zhenyu and Shi, Guang and Feng, Jiashi and Kang, Bingyi},
  journal={arXiv preprint arXiv:2511.10647},
  year={2025}
}

@inproceedings{wang2024dust3r,
  title={Dust3r: Geometric 3d vision made easy},
  author={Wang, Shuzhe and Leroy, Vincent and Cabon, Yohann and Chidlovskii, Boris and Revaud, Jerome},
  booktitle={2024 IEEE/CVF Conference on Computer Vision and Pattern Recognition (CVPR)},
  pages={20697--20709},
  year={2024},
  organization={IEEE}
}

@inproceedings{triggs1999bundle,
  title={Bundle adjustment—a modern synthesis},
  author={Triggs, Bill and McLauchlan, Philip F and Hartley, Richard I and Fitzgibbon, Andrew W},
  booktitle={International workshop on vision algorithms},
  pages={298--372},
  year={1999},
  organization={Springer}
}

@inproceedings{schonberger2016structure,
  title={Structure-from-motion revisited},
  author={Schonberger, Johannes L and Frahm, Jan-Michael},
  booktitle={Proceedings of the IEEE conference on computer vision and pattern recognition},
  pages={4104--4113},
  year={2016}
}

@article{mur2015orb,
  title={ORB-SLAM: A versatile and accurate monocular SLAM system},
  author={Mur-Artal, Raul and Montiel, Jose Maria Martinez and Tardos, Juan D},
  journal={IEEE transactions on robotics},
  volume={31},
  number={5},
  pages={1147--1163},
  year={2015},
  publisher={IEEE}
}

@inproceedings{yang2025fast3r,
  title={Fast3r: Towards 3d reconstruction of 1000+ images in one forward pass},
  author={Yang, Jianing and Sax, Alexander and Liang, Kevin J and Henaff, Mikael and Tang, Hao and Cao, Ang and Chai, Joyce and Meier, Franziska and Feiszli, Matt},
  booktitle={2025 IEEE/CVF Conference on Computer Vision and Pattern Recognition (CVPR)},
  pages={21924--21935},
  year={2025},
  organization={IEEE}
}

@inproceedings{cabon2025must3r,
  title={Must3r: Multi-view network for stereo 3d reconstruction},
  author={Cabon, Yohann and Stoffl, Lucas and Antsfeld, Leonid and Csurka, Gabriela and Chidlovskii, Boris and Revaud, Jerome and Leroy, Vincent},
  booktitle={2025 IEEE/CVF Conference on Computer Vision and Pattern Recognition (CVPR)},
  pages={1050--1060},
  year={2025},
  organization={IEEE}
}

@inproceedings{zhang2026efficiently,
  title={Efficiently reconstructing dynamic scenes one d4rt at a time},
  author={Zhang, Chuhan and Le Moing, Guillaume and Koppula, Skanda and Rocco, Ignacio and Momeni, Liliane and Xie, Junyu and Sun, Shuyang and Sukthankar, Rahul and Barral, Jo{\"e}lle K and Hadsell, Raia and others},
  booktitle={Proceedings of the IEEE/CVF Conference on Computer Vision and Pattern Recognition},
  pages={7382--7392},
  year={2026}
}

@inproceedings{karhade2026any4d,
  title={Any4d: Unified feed-forward metric 4d reconstruction},
  author={Karhade, Jay and Keetha, Nikhil and Zhang, Yuchen and Gupta, Tanisha and Sharma, Akash and Scherer, Sebastian and Ramanan, Deva},
  booktitle={Proceedings of the IEEE/CVF Conference on Computer Vision and Pattern Recognition},
  pages={14578--14589},
  year={2026}
}

@inproceedings{zhou2026page,
  title={Page-4d: Disentangled pose and geometry estimation for vggt-4d perception},
  author={Zhou, Kaichen and Wang, Yuhan and Chen, Grace and Beaudouin, Gaspard and Zhan, Fangneng and Liang, Paul and Wang, Mengyu},
  booktitle={International Conference on Learning Representations},
  volume={2026},
  pages={36401--36414},
  year={2026}
}

@article{fang2026more,
  title={More: Motion-aware feed-forward 4d reconstruction transformer},
  author={Fang, Juntong and Chen, Zequn and Zhang, Weiqi and Di, Donglin and Zhang, Xuancheng and Yang, Chengmin and Liu, Yu-Shen},
  journal={arXiv preprint arXiv:2603.05078},
  year={2026}
}

@inproceedings{zhuo2026streaming,
  title={Streaming visual geometry transformer},
  author={Zhuo, Dong and Zheng, Wenzhao and Guo, Jiahe and Wu, Yuqi and Zhou, Jie and Lu, Jiwen},
  booktitle={International Conference on Learning Representations},
  volume={2026},
  pages={88055--88072},
  year={2026}
}

@article{yuan2026infinitevggt,
  title={Infinitevggt: Visual geometry grounded transformer for endless streams},
  author={Yuan, Shuai and Yang, Yantai and Yang, Xiaotian and Zhang, Xupeng and Zhao, Zhonghao and Zhang, Lingming and Zhang, Zhipeng},
  journal={arXiv preprint arXiv:2601.02281},
  year={2026}
}

@inproceedings{keetha2026mapanything,
  title={Mapanything: Universal feed-forward metric 3d reconstruction},
  author={Keetha, Nikhil and M{\"u}ller, Norman and Sch{\"o}nberger, Johannes and Porzi, Lorenzo and Zhang, Yuchen and Fischer, Tobias and Knapitsch, Arno and Zauss, Duncan and Weber, Ethan and Antunes, Nelson and others},
  booktitle={2026 International Conference on 3D Vision (3DV)},
  pages={499--509},
  year={2026},
  organization={IEEE}
}

@inproceedings{wang2026flashvggt,
  title={Flashvggt: Efficient and scalable visual geometry transformers with compressed descriptor attention},
  author={Wang, Zipeng and Xu, Dan},
  booktitle={Proceedings of the IEEE/CVF Conference on Computer Vision and Pattern Recognition},
  pages={21826--21835},
  year={2026}
}

@article{deng2026saf3r,
  title={{SAF3R}: Dynamic sparse attention for feed-forward {3D} reconstruction transformers},
  author={Deng, Jianing and Li, Yuanzhe and Wang, Jialu and Wang, Song and Chen, Tianlong and Yang, Huanrui and Hu, Jingtong},
  journal={arXiv preprint arXiv:2607.03612},
  year={2026}
}

@article{elflein2026vgg,
  title   = {{VGG-T}$^3$: Offline Feed-Forward 3D Reconstruction at Scale},
  author  = {Elflein, Sven and Li, Ruilong and Agostinho, S{\'e}rgio and Gojcic, Zan and Leal-Taix{\'e}, Laura and Zhou, Qunjie and Osep, Aljosa},
  journal = {arXiv preprint arXiv:2602.23361},
  year    = {2026}
}

@inproceedings{jin2026zipmap,
  title={Zipmap: Linear-time stateful 3d reconstruction via test-time training},
  author={Jin, Haian and Wu, Rundi and Zhang, Tianyuan and Gao, Ruiqi and Barron, Jonathan T and Snavely, Noah and Ho{\l}y{\'n}ski, Aleksander},
  booktitle={Proceedings of the IEEE/CVF Conference on Computer Vision and Pattern Recognition},
  pages={21748--21759},
  year={2026}
}

@article{xie2026scal3r,
  title={Scal3r: Scalable test-time training for large-scale 3d reconstruction},
  author={Xie, Tao and Yang, Peishan and Jin, Yudong and Cai, Yingfeng and Yin, Wei and Ren, Weiqiang and Zhang, Qian and Hua, Wei and Peng, Sida and Guo, Xiaoyang and others},
  journal={arXiv preprint arXiv:2604.08542},
  year={2026}
}

@article{you2026regimevggt,
  title={RegimeVGGT: Layer-Wise Spatially Preserving Redundancy Removal for Visual Geometry Grounded Transformer},
  author={You, Jinhao and Lyu, Shuo and Lyu, Zhuohang and Li, Tanxuan and Zhao, Zibo and Hu, Jiaxiang and Tang, Kai and Guo, Yichen},
  journal={arXiv preprint arXiv:2606.18439},
  year={2026}
}

@article{tang2026spark3r,
  title={Spark3R: Asymmetric Token Reduction Makes Fast Feed-Forward 3D Reconstruction},
  author={Tang, Zecheng and Fu, Jiaye and Gao, Qiankun and Li, Haijie and Wu, Yanmin and Zhang, Jiaqi and Ma, Siwei and Zhang, Jian},
  journal={arXiv preprint arXiv:2605.06270},
  year={2026}
}

@inproceedings{gupta2021value,
  title={Value-aware approximate attention},
  author={Gupta, Ankit and Berant, Jonathan},
  booktitle={Proceedings of the 2021 Conference on Empirical Methods in Natural Language Processing},
  pages={9567--9574},
  year={2021}
}

@article{mur2017orb,
  title={Orb-slam2: An open-source slam system for monocular, stereo, and rgb-d cameras},
  author={Mur-Artal, Raul and Tard{\'o}s, Juan D},
  journal={IEEE transactions on robotics},
  volume={33},
  number={5},
  pages={1255--1262},
  year={2017},
  publisher={IEEE}
}

@inproceedings{klein2007parallel,
  title={Parallel tracking and mapping for small AR workspaces},
  author={Klein, Georg and Murray, David},
  booktitle={2007 6th IEEE and ACM international symposium on mixed and augmented reality},
  pages={225--234},
  year={2007},
  organization={IEEE}
}

@inproceedings{zarch2026delta,
  title={DELTA: Dynamic Layer-Aware Token Attention for Efficient Long-Context Reasoning},
  author={Zarch, Hossein Entezari and Gao, Lei and Jiang, Chaoyi and Annavaram, Murali},
  booktitle={Findings of the Association for Computational Linguistics: ACL 2026},
  pages={11502--11518},
  year={2026}
}

@article{liao2024beyond,
  title={Beyond kv caching: Shared attention for efficient llms},
  author={Liao, Bingli and Vargas, Danilo Vasconcellos},
  journal={arXiv preprint arXiv:2407.12866},
  year={2024}
}

@article{kim2026hess,
  title={HeSS: Head sensitivity score for sparsity redistribution in VGGT},
  author={Kim, Yongsung and Song, Wooseok and Lew, Jaihyun and Hwangbo, Hun and Lee, Jaehoon and Yoon, Sungroh},
  journal={arXiv preprint arXiv:2603.25336},
  year={2026}
}

@article{ren2026speed3r,
  title={Speed3r: Sparse feed-forward 3d reconstruction models},
  author={Ren, Weining and Tan, Xiao and Han, Kai},
  journal={arXiv preprint arXiv:2603.08055},
  year={2026}
}

@article{kulis2011revisiting,
  title={Revisiting k-means: New algorithms via Bayesian nonparametrics},
  author={Kulis, Brian and Jordan, Michael I},
  journal={arXiv preprint arXiv:1111.0352},
  year={2011}
}

@article{guo2026phaselora,
  title={{PhaseLoRA}: Control-Regime-Conditioned Low-Rank Adaptation for Continuous-Action Vision-Language-Action Policies},
  author={Guo, Yufei and Wu, Yinan and Duan, Haoran and Ding, Guiguang and Han, Jungong},
  journal={arXiv preprint arXiv:2608.15285},
  year={2026}
}
